**ABSTRACT**

**USING APPROXIMATE MODELS IN ROBOT LEARNING**

Ali Lenjani, MS

Department of Mechanical Engineering

Northern Illinois University, 2016

Brianno Coller, Director

Trajectory following is one of the complicated control problems when its dynamics are nonlinear, stochastic and include large number of parameters. The problem has major difficulties including large number of trials required for data collection and huge volume of computations required to find a closed-loop controller for high dimensional and stochastic domains. For solving this type of problem, if we have an appropriate reward function and dynamics model; finding an optimal control policy is possible by using model-based reinforcement learning and optimal control algorithms. However, defining an accurate dynamics model is not possible for complicated problems. Pieter Abbeel and Andrew Ng recently presented an algorithm that requires only an approximate model and only a small number of real-life trials. This algorithm has wide applicability; however, there are some problems regarding convergence of the algorithm. In this research, required modifications are presented that provide more powerful assurance for converging to an optimal control policy. Also updated algorithm is implemented to evaluate the efficiency of the new algorithm by comparing the acquired results with human expert performance. We are using differential dynamic programming (DDP) as the locally trajectory optimizer and a 2D dynamics and kinematics simulator is used to evaluate the accuracy of the presented algorithm.

NORTHERN ILLINOIS UNIVERSITY

DE KALB, ILLINOIS

DECEMBER 2016

USING APPROXIMATE MODELS IN ROBOT LEARNING

BY

ALI LENJANI

A THESIS SUBMITTED TO THE GRADUATE SCHOOL

IN PARTIAL FULFILLMENT OF THE REQUIREMENTS

FOR THE DEGREE

MASTER OF SCIENCE

DEPARTMENT OF MECHANICAL ENGINEERING

Thesis Director:

Professor Brianno Coller

# ACKNOWLEDGEMENTS

I would like to express my sincere gratitude to my advisor, Professor Brianno Coller, for his continuous support, motivation, patience and generously sharing knowledge. He was selected as the best professor of the year, but in my point of view he is the best professor ever. Beside all his academic aspect, during my interaction with Professor Coller I also learned indirectly how to behave with other people, and because of this, he is one of the most important persons in my life.

TABLE OF CONTENTS

# LIST OF FIGURES

# CHAPTER 1. INTRODUCTION

Undoubtedly interacting with our environment is the most important element when we think about nature of learning. When a boy plays, without having an explicit teacher he has a direct connection to his environment. This interaction provides a huge amount of information about cause and effect, about the values and costs of actions, and about what to do in order to achieve goals. Throughout our lives, such interactions are undoubtedly a major source of knowledge about our environment and us. Whether we are learning to drive a car or to play a sport, we are deeply aware of how our environment responds to what we do, and we try to understand what happens through our behavior. Learning from interaction is a first idea of almost all theories of learning and intelligence[1], [2].

Flight control systems for aircraft, automated manufacturing systems, and sophisticated avionics systems all present difficult, nonlinear control problems. Many of these problems are currently unsolvable, not only because current computers are too slow or have too little memory, but also because it is too difficult to determine what the program should do. Combination of computer capabilities and human ability to learn from trial and error would be a possible solution for this kind of problems[3].

Traditionally, it has often been assumed in robotics research that perfect knowledge about the robot and its environment is available. But most times we have limited access to accurate models of both the environment and the robot. Limitations may have different causes, such as lack of knowledge, engineering limits, tractability bottleneck and precision. Accurate models of the robot and its environment should be provided by a human engineer, and even if appropriately detailed knowledge is available, making it computer accessible, hand coding explicit models of robot hardware, sensors and environments has often been found to require unreasonable amounts of programming time. Most realistic robot domains are too complex to be handled efficiently[4]. Computational tractability is a strong difficulty for designing control structures for complex robots in complex domains, and robots are far from being "reactive." Also the robot device must be precise enough to accurately execute plans that were generated using the internal models of the world.

Autonomous robotics research has changed the design of autonomous agents and provide an appropriate way for future research in robotics[5], [6]. Reactivity and real-time operation have received noticeably more attention than, for example, optimality and completeness. Many approaches are based on the assumption that perfect environment knowledge is available; however, some systems even operate in the extreme where no domain-specific initial knowledge is available at all. So, today's robots are facing unknown and possibly aggressive environments. In this regard, they have to adjust themselves, explore their environments autonomously, recover from failures, and due to application they should do different tasks.

When robots face lack of initial knowledge about themselves and their environments, learning would be the most reliable solution. Learning refers to variety of algorithms that are categorized by their ability to replace incorrect environment knowledge by trial and error, observation, and generalization. Robots gather parts of their knowledge themselves and recover it during time. Their dependency on a human decreased to provide this knowledge beforehand.

Learning robots are usually flexible enough to deal with a various type of environments and tasks. Consequently, the internal initial knowledge, if available at all, is often too weak to solve an existing problem off-line.

Learning robots rely on the interaction with their environment to achieve required information. Different learning strategies differ mainly in three aspects: the method to do exploration, method of generalization from the observed experience, and the type and amount of initial knowledge that constrains the space of their internal knowledge about the environment. It is not possible to find an optimal general learning technique for autonomous robots since learning techniques are considered by a trade-off between the degree of flexibility, domain of initial knowledge, and the amount of observations required for filling missing or incorrect initial knowledge. Roughly speaking, the more general a robot learning architecture, the more experiments the robot needs to take to learn sufficiently. Two of the most important benefits of learning strategies to autonomous robot agents are their capability to operate in a various class of initially unknown environments and their ability to compensate for changes, since they update their knowledge about the environment and themselves. Moreover, all learned knowledge is implemented in the real-life environment. Although learning has long been introduced in AI as key technique to make autonomous robots able to solve more complicated tasks in more realistic environments, recent research has produced a variety of developed learning techniques that allow a robot to obtain enormous amounts of knowledge by itself(Figure 1)[7]–[9].

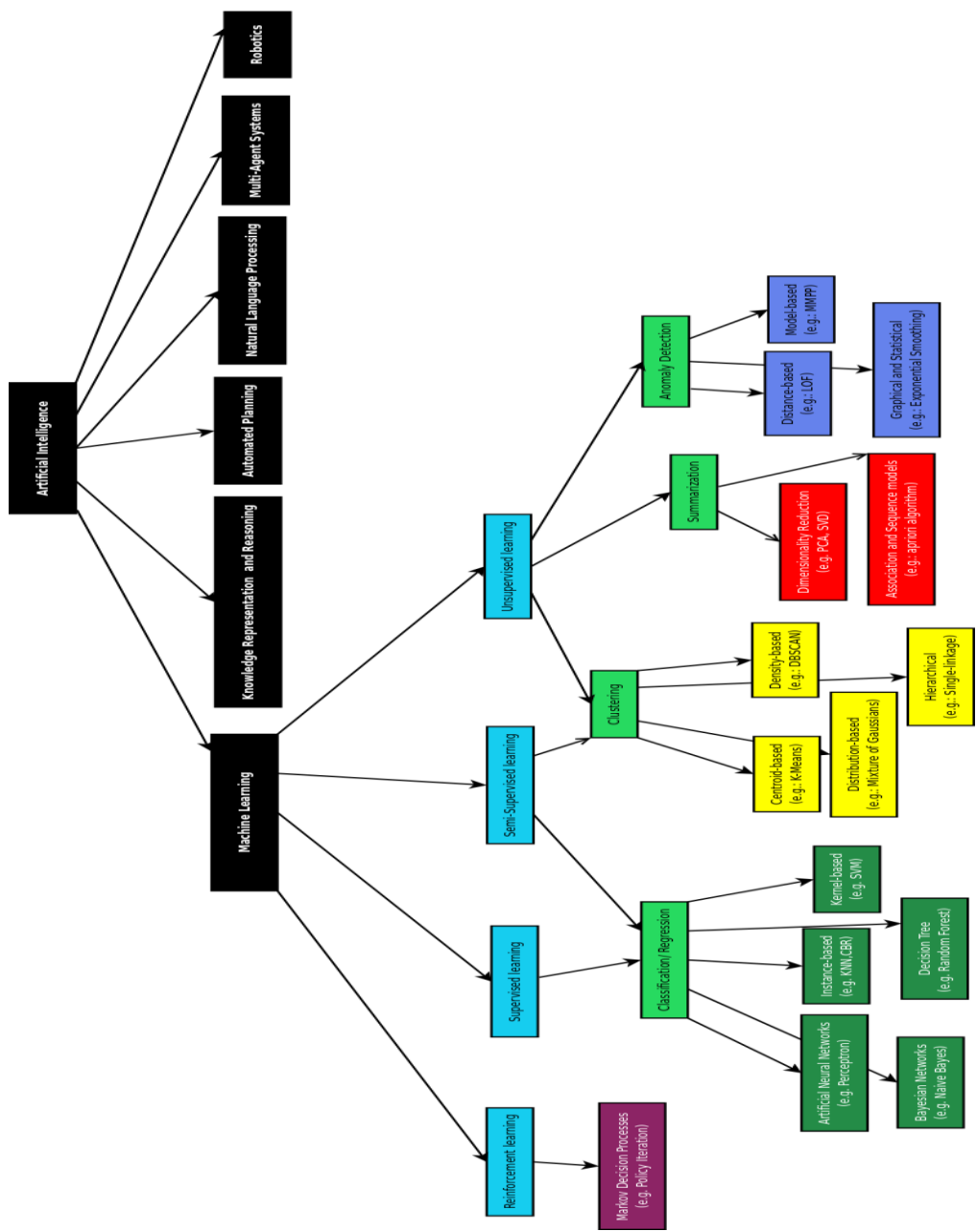

**Figure 1: AI categories**

## Problem Description and Research Objective

Whereas a perfect and exact dynamics model will improve the controllers found by model-based reinforcement learning (RL) algorithms, a completely precise dynamics model is not always necessary to calculate good controllers. We are developing this idea and improve an algorithm presented by Pieter Abbeel and Andrew Ng that uses a crude model to learn rapidly to perform well on real systems[10].

For assessing the algorithm we are using a 2D dynamics simulator called Spumone. However we do not model the exact gravity in our calculations to derive the initial dynamics (approximate dynamics) of the system; we want to run the craft from the start point to destination without colliding obstacles. We can also consider reasonable limitations on velocity or acceleration due to application.

The main idea is using a real-life experiment to evaluate a calculated control policy and then use the simulator (or model) to calculate the derivative of the evaluation with respect to the control policy parameters and propose local improvements. For example, in Spumone, if our current policy drives too far to the left, then driving in real system will demonstrate that we are driving too far to the left by calculating difference of trajectories in each time step. Here we update our dynamics model due to difference of trajectory states in our simulation and the real system. However, even a very poor model of the craft can then be used to tell us that the change we should make is to apply a clockwise torque (rather than anti-clockwise) to correct for this error. In particular, we do not need additional trials of applying both clockwise and anti-clockwise torque in order to decide which direction to turn it. After finding the direction of torque, the amount of required torque clockwise can be determined by doing a line search and test each control policy in real system to find the policy that will perform better. Therefore, even a crude model of the craft allows us to significantly reduce the number of real-life trials needed compared to model-free algorithms that operate directly on the real system.

## 1.2 Organization

The thesis is organized as follows:

Chapter 2 reviews the literature of reinforcement learning and apprenticeship learning (learning from expert demonstration). Chapter 3 describes the mathematical model for the problem and some mathematical preliminaries for the solution approach. In Chapter 4, we present an algorithm that requires only an approximate model (of a possibly complex system) and only a small number of real-life trials. Chapter 5 presents the experimental study conducted in this research and Chapter 6 concludes the research as well as proposes recommendations for future research.

# CHAPTER 2. LITERATURE REVIEW

## 2.1 Introduction

Reinforcement learning is an approach to machine learning that combines two different approaches to solve problems that neither can individually solve. One of those is *dynamic programming,* a field of mathematics that has been used for a long time to solve optimization and control problems. But traditional dynamic programming has limitation due to size and complexity of the problems that it can solve.

On the other side for training a parameterized function approximation, such as a neural network, *supervised learning* is a popular method used to represent functions. But supervised learning needs sample input-output sets of data to learn the function. Roughly speaking, supervised learning needs a set of questions and the correct answers. For instance, to recognize a picture of a tree, we have a large collection of pictures, and the information is provided already whether there is a tree in each picture or not. Supervised learning could search all the samples and correct answers and learn how to distinguish a picture of a tree in general.

In some situations, the right answer which supervised learning requires is not known to us. For instance, in a flight control system, the question would be all sensor readings, and the answer is the required action to move the control surfaces. Flying the plane using neural networks is possible only when there is a set of correct answers; therefore, if method of building a controller is not known at first, ordinary supervised learning would not be helpful.

Because of the mentioned reasons there has been much attention recently in a different approach called *reinforcement learning* (RL). Reinforcement learning is not a kind of neural network and it is not also an alternative for neural networks. Reinforcement learning combines the fields of dynamic programming and supervised learning to produce great machine learning systems. Reinforcement learning is popular because of its generality. First, a goal is defining for the computer to attain. Then it learns how to do that goal by trial-and-error interactions with the environment. Therefore, many researchers are following this method of machine intelligence and are motivated about the chance of solving problems that were unsolvable previously[2], [11].

To understand the key idea of reinforcement learning, consider the problem of learning how to ride a bicycle. The objective of the RL system is to ride a bicycle correctly. In the first test, the RL system begins riding the bicycle and executes a sequence of actions that consequence the bicycle being in 45 degrees left. At this time we can choose two possible actions: turn the handle left or right. If RL system turns the handle to the right it crashes, consequently receiving a negative reward. The RL system learns should not turn the handle right when it is in 45 degrees to the left. In the next test the RL system executes a sequence of actions that again result in the bicycle being tilted 45 degrees to the left. The RL system recognizes not to turn the handle to the right, so it executes the other possible action that is turning right. It crashes again and gets a strong negative reward. At the moment the RL system is learning that in addition to turning the handle right or left while slanted 45 degrees to the right is bad, the "state" of being slanted 45 degrees to the right is bad. Then the RL system starts another test and executes a sequence of actions that result in the bicycle being slanted 40 degrees to the left.

Again there are two possible actions: turn right or turn left. The RL system turns the handle right, which results in the bicycle being slanted 45 degrees to the left and finally results in a big negative reward. The RL system is learning not to turn the handle right when slanted 40 degrees to the left. By execution of adequate trial-and-error interactions with the environment, the RL system will learn to prevent the bicycle from falling [3].

## 2.2 Reinforcement Learning Problem

In the typical reinforcement learning, an agent interacts with its environment. This interaction results in sensing the environment by agent and based on this input selecting an action to execute in the environment. The action modifies the environment in some way and this modification is connected to the agent through a scalar reinforcement signal. Beside the agent and the environment, four main elements of a reinforcement learning system are policy, reward function, value function, and model of the environment[11].

### 2.2.1   Policy

A policy describes the agent's way of acting. A policy is a mapping from observed states of the environment to actions taken when in those states. Sometimes the policy could be a simple function or a lookup table, while in others it may include wide computation like a search process. The policy is the core of a reinforcement learning agent in the sense that it alone is sufficient to determine behavior.

### 2.2.2   Reward Function

A reward function describes the objective in a reinforcement learning problem. It maps each observed state or state-action pair of the environment to a number. A reward represents the inherent desirability of that state. An agent's objective is to maximize the total reward it obtains in the long run. The reward function describes good and bad events for the agent. Reward function need essentially be unchangeable by the agent. It may, however, work as a foundation for changing the policy.

### 2.2.3   Value Function

While a reward function specifies what is appropriate in an immediate sense, a value function identifies what is appropriate in the long run. The value of a state is the total reward an agent can imagine to store over the future, starting from that state. For instance, a state may continuously return a low reward but have a high value because it is followed by other states that return high rewards. Rewards are in an immediate sense, while values, is predictions of rewards. Without rewards there can be no values, and the objective of estimating values is to attain more reward [2].

### 2.2.4   Model

The fourth part of reinforcement learning systems would be model of the environment. This is something that simulates the performance of the environment. For instance, by giving a state and action, the model can calculate the subsequent state and reward. Models are used for preparation; it means deciding on a sequence of action by seeing possible future states before they are really experienced[2].

## 2.3 Approximating the Value Function

Reinforcement learning is a challenging problem since the learning system might take an action and it is not obvious that is good or bad. For instance, a learning flight controller program goal is not to crash. It should make numerous decisions each second and then, after performing many actions, the airplane may crash. What does the system learn from this experience? Which of these actions is guilty of the crash? Assigning responsibility to single actions is the challenge that makes reinforcement learning difficult. Fortunately, there is solution to this difficulty that is based on an area of mathematics called dynamic programming, and it includes only two elementary principles. First, if an action results in bad happening immediately, like crashing the plane, then the system understands not to take that action in same situation again. Therefore, the action system executed one time step before the crash will be avoided in the future. But that principle doesn't work for all the earlier actions that didn't result in immediate bad happening.

The second and more interesting one is that if all the actions in specific state results in bad consequences, then that state should be avoided. Thus if the system has experienced a specific combination of altitude and wind speed, executing a different action each time, and all actions result in bad happening, so it learns that this state itself is bad. This is a great principle because the system learns without crashing. In the future, if it selects an action that results in this specific state, it immediately learns that specific action is bad, without waiting for the crash.

By using these two principles, the system learns to choose required decision to fly an airplane, follow a trajectory or any other tasks. It first learns on a simulator and then adjusts on the real system.

## 2.4 Miscellaneous Issues

### 2.4.1 Exploration

As mentioned before, the central inquiry in reinforcement learning exploration is: How would we develop an algorithm that will effectively locate the ideal value function? It was demonstrated that the ideal value function is an answer for the arrangement of conditions characterized by the Bellman equation. The process of learning was subsequently described as the process of developing an approximation of the optimal value function by incrementally finding a solution for this problem. One ought to notice that the Bellman equation is characterized over all of state space. The optimal value function satisfies this equation for all $x_t$ in state space.

This requirement presents the necessity for exploration. Exploration is described as purposefully selecting to execute an action that is not the best option, for the purpose of obtaining knowledge of unobserved states. For recognizing an optimal approximation, state space should be adequately explored.

For instance, a robot facing an unobserved environment has to spend time obtaining knowledge of its environment. On the other hand, experience learned during exploration should also be considered during choosing action to minimize the costs of learning. While the robot should explore its environment, it should avoid hitting the obstacles. But the robot does not know that which actions will lead to collision till whole state space has been explored. Alternatively, it is also possible that a policy that is "adequately" good will be found without exploring all of state space.

There is an important trade-off between exploration and exploitation (using learned knowledge for choosing action). Consequently, it is vital to use exploration methods that maximize the knowledge expanded during learning and minimize the costs of exploration and time for learning.

### 2.4.2 Discounted Factor

The discount factor $\gamma$ is a float number in between 0 to 1 and is used to set the weight closer for reward than far rewards. The closer $\gamma$ is to 0 the smaller the weight of future reinforcements. For $\gamma = 0$, the value of a state is based completely on the immediate reward received for execution of associated action. For finite horizon Markov decision processes (an MDP that terminates) it is not strongly necessary to use a discount factor. In this case ($\gamma = 1$), the value of state $x_t$ is based on the total reinforcement received when starting in state $x_t$ and following the given policy:

$$\Delta w_t = \max\big(R(x_t, u) + \gamma \times V(x_{t+1}, w_t)\big) - V(x_t, w_t) \qquad 1$$

In infinite horizon Markov decision processes (an MDP that never terminates), a discount factor is essential. Without using a discount factor, the sum of the rewards achieved would be infinite for each state. Using a discount factor bounds the maximum value of a state to be on the order of:

$$R/1 - \gamma.$$

## 2.5 Learning from Demonstration

### 2.5.1 Overview

While applying conventional control or reinforcement learning techniques, one of the essential difficulties is giving a formal particular of the control task. The ideal control and reinforcement learning formalisms require the detail of a prize (or cost) function characterizing "goodness" of every conceivable state. This prize function is frequently difficult to indicate. For instance, what might be the right reward capacity for "driving great"? In this chapter, we depict how one can influence master exhibitions to productively address this test. Specifically, we consider learning in a Markov choice procedure where we are not unequivocally given a prize function, but rather where we can watch a specialist showing the task that we need to figure out how to perform. We think about the master as attempting to augment a prize function that is expressible as a straight blend of known elements and give an algorithm for taking in the assignment shown by the master. Our algorithm depends on utilizing "opposite reinforcement learning" to attempt to recuperate the obscure prize function[12], [13].

Robot learning from demonstration (also called apprenticeship learning) is authorizing robots to self-sufficiently perform new tasks. Instead of obliging users to scientifically break down and physically program a wanted conduct, work in learning from demonstration takes the perspective that a fitting robot controller can be gotten from perceptions of a human's own particular execution thereof. The point is for robot capacities to be all the more effectively stretched out and adjusted to novel circumstances, even by users without programming ability[14].

The primary guideline of robot learning from demonstration is that end users can show robots new assignments without programming. Consider, for instance, a household administration robot that a proprietor wishes to get ready squeezed juice for breakfast. The task itself may include different subtasks, for example, squeezing the orange, tossing whatever is left of the orange in the rubbish and pouring the fluid in a glass. Besides, every time this errand is played out, the robot should fight with changes. In a customary programming situation, a human developer would need to reason ahead of time and code a robot controller that is equipped for reacting to any circumstance the robot may confront, regardless of how improbable. This procedure may include separating the task into hundreds of various strides and completely testing every progression.

On the off chance that mistakes or new circumstances emerge after the robot is deployed, maybe the whole excessive procedure ought to be repeated and the robot reviewed or taken out of administration while it is fixed. In contrast, learning from demonstration allows the end user to "program" the robot simply by showing it how to perform the task - no coding required. At that point, when disappointments happen, the end-user needs just to give more exhibitions, as opposed to calling for expert help. Learning from demonstration hence seeks to endow robots with the ability to learn what it means to perform a task by generalizing from observing several demonstrations. Learning from demonstration is not a record-and-replay technique. Learning and generalizing are core learning from demonstration [13], [14].

Robot learning from demonstration began in the 1980s. At that point, and still to a substantial degree now, robots had to be monotonously hand modified for each undertaking they performed. Learning from demonstration looks to minimize, or even dispense with, this troublesome stride by giving users a chance to prepare their robot to fit their needs. The desire is that the techniques for learning from demonstration, being easy to use, will permit robots to be used to a more prominent degree in everyday collaborations with non-specialist humans.

Besides, by using master information from the user, as exhibitions, the actual learning ought to be quick compared to current trial-and-error learning, especially in high dimensional spaces (consequently tending to part of the surely understood condemnation of dimensionality).

Research on learning from demonstration has become relentless in significance since the 19s and a few reviews have been distributed as of late. Most work on learning from demonstration follows a more engineering/machine learning approach. Reviews of works here incorporate [15]. At the center, be that as it may, learning from demonstration is propelled by the way people gain from being guided by specialists, from early stages through adulthood.

A substantial collection of work on learning from demonstration in this manner takes motivation from ideas in psychology and biology. Some of these works seek after a computational neuroscience approach and utilize neural displaying. Others seek after a more intellectual science approach and construct calculated model of impersonation learning in creatures[15], [16].

## 2.5.2  Key Issues in Learning from Demonstration

What to imitate identifies with the issue of figuring out which parts of the exhibit ought to be imitated. For a given assignment, certain detectable or effectible properties of the environment might be immaterial and securely disregarded. Key to figuring out what is and is not imperative is understanding the metric by which the robot's conduct is being assessed[17].

Another method for saying this is the metric used to figure out whether the robots have effectively played out the wanted task includes just the extent of the cases, yet not their shading. In this manner, the robots figure out how to overlook shading in their endeavors. Instructing what is and is not imperative should be possible in different ways. The simplest approach is to take a statistical perspective and deem as relevant the parts (dimension, region of input space, etc.) of the data which are consistent across all demonstrations. On the off chance that the measurement of the information is too high, such a methodology may require an excessive number of exhibitions to accumulate enough insights. An option is then to have the instructor help the robot figure out what is pertinent by indicating out parts of the undertaking that are generally essential[17], [18].

How to imitate consists in determining how the robot will actually perform the learned behaviors to maximize the metric found when solving the what to imitate problem. Often, a robot cannot act exactly the same way as a human does due to differences in physical embodiment. For example, if the demonstrator uses a foot to move an object, is it acceptable for a wheeled robot to bump it, or should it use a gripper instead? If the metric does not have appendage-specific terms, it may not matter.

This issue is firmly identified with that of the "Correspondence Problem". Robots and people, while occupying the same space and communicating with the same articles, and maybe even externally comparative, still see and associate with the world in on a very basic level diverse ways. To assess the likeness between the human and robot practices, we should first manage the way that the human and the robot may involve distinctive state spaces of maybe diverse dimensions.

We distinguish two distinctive routes in which conditions of demonstrator and imitator can be said to relate and give brief instances.

1. **Perceptual equivalence:** Because of contrasts among human and robot tactile abilities, the same scene may seem altogether different to each. Case in point, while a human may distinguish people and signals from light, a robot may utilize profundity estimations to watch the same scene. Another purpose of examination is material detecting.

   Most material sensors permit robots to see contact yet don't offer data about temperature, as opposed to the human skin. Additionally, the low determination of the robots material sensors does not permit robots to separate over the assortment of existing surfaces, while human skin does.

   As the same information may hence not be accessible to both people and robots, effectively educating a robot may require a decent comprehension of the robot's sensors and their impediments. Learning from demonstration investigates the points of confinement of these perceptual equivalences by building interfaces that either naturally rectify for or make express these distinctions.

2. **Physical equivalence:** Because of contrasts among human and robot exemplifications, people and robots may perform diverse activities to finish the same physical impact. Case in point, notwithstanding when playing out the same assignment (football), people and robots may communicate with the environment in various ways. Here the people run and kick while the robots roll and knock.

Understanding this disparity in engine capacities is likened to tackling the how to imitate problem and is the center of much work in learning from demonstration. In the football example above, this would need the robot to define a path for its center of mass which matches to the path tracked by the human's right foot when projected on the ground. Clearly, this equivalence is very task dependent. Recent solutions to this problem are for hand motion and body motion[19].

Taken together, these two equivalences manage disparities in how robots and people are encapsulated. We can think about the perceptual equality as managing the way in which the agents see the world and ensure that the data important to play out the undertaking is accessible to both. Physical identicalness manages the way in which agents influence and interface with the world and ensures that the undertaking is really performable by both.

The interface used to give showings assumes a key part in the way the data is accumulated and transmitted. We recognize three noteworthy patterns:

Directly recording human movements. At the point when interested only in the kinematics of the movements, one may utilize any of different existing movements following frameworks, in view of vision, exoskeleton or other wearable movement sensors. The movement of the human body is initially separated from the foundation utilizing a model of human body and is then mapped to a symbol and the humanoid robot DB at ATR, Kyoto, Japan.

Kinesthetic instructing. The robot is physically guided through the undertaking by the people. With this approach, no express physical correspondence is required, as the user exhibits the ability with the robot's own body. It likewise gives a characteristic instructing interface to amend an ability replicated by the robot. In the last mentioned, skin innovation is utilized to figure out how touch contacts identify with the current workload, raising issues of how to separate between touches that are a piece of the task and those that are a piece of the educating[20].

Immersive teleoperation situations. A human administrator is constrained to utilizing the robot's own particular sensors and effectors to play out the task. Besides kinesthetic instructing, which restrains the user to the robot's own particular body, immersive teleoperation looks as far as possible from the user's judgment to those of the robot. The teleoperation itself might be done utilizing joysticks or other remote-control devices, including haptic gadgets. The latter has the favorable position that it can permit the educator to instruct tasks that require exact control of strengths, whereas joysticks would just give kinematic data (position, speed)[21-24].

Current ways to deal with encoding aptitudes through learning from demonstration can be extensively partitioned between two patterns: a low-level representation of the expertise, taking the type of a nonlinear mapping amongst tangible and motor information, and a high-level state representation of the ability that deteriorates the aptitude in a grouping of action discernment units.

Singular movements/activities could be taught independently rather than at the same time. The human instructor would then give one or more cases of every sub-movement separated from the others. On the off chance that taking in continues from the perception of a solitary occurrence of the movement/activity, one calls this one-shot learning[25]. Cases can be found for learning locomotion patterns. Unique in relation to straightforward record and play, here the controller is furnished with earlier information as primitive movement designs and takes in parameters for these examples from the exhibit[21].

Multi-shot learning can be performed in a clump subsequent to recording a few exhibitions or incrementally as new shows are performed. Taking in for the most part performs derivation from factual examination of the information crosswise over exhibits, where the signs are demonstrated by means of a likelihood thickness work and broke down with different nonlinear regression techniques stemming from machine learning. Well-known techniques nowadays incorporate Gaussian process, Gaussian mixture models, super vector machines[22].

While most learning from demonstration work to date has concentrated on learning kinematic movements of end effectors or different joints, later work has explored removing force-based signs from human demonstration[21]. Transmitting data about power is troublesome for people and for robots alike, since power can be detected just when playing out the task ourselves. Current endeavors in this way try to decouple the teaching of kinematics and constrain or create techniques by which one may "embody" the robot and, by so doing, permit human and robot to see all the while the powers connected when playing out the assignment. This profession is powered by late advances in the outline of haptic devices and material detecting and on the improvement of torque and variable-impedance-incited frameworks to show force-control errands through human exhibit.

Learning complex tasks, made out of a mix and juxtaposition of individual movements, is a definitive objective of learning from demonstration. A typical methodology is to first learn models of the greater part of the individual movements, utilizing exhibits of each of these activities individually, and after that take in the privilege sequencing mix in a brief moment organized either by watching a human playing out the entire task or through reinforcement learning. Notwithstanding, this methodology assumes that there is a known arrangement of all essential primitive activities. For particular undertakings this might be valid, but to date there does not exist a database of universally useful primitive actions, and it is vague if the variability of human movement may truly be decreased to a limited rundown[23].

An option is to watch the human play out the complete task and to naturally portion the task to extricate the primitive actions (which may then get to be task dependent). The fundamental preferred standpoint is that both the primitive actions and the way they ought to be joined are learned in one pass. One issue that emerges is that the quantity of primitive tasks is frequently obscure, and there could be different conceivable divisions which must be considered[24].

Complex tasks are made out of compound activities - a robot stacking dishes into a dishwasher. In an outline of the principal approach, the robot is given an arrangement of known (pre-customized or adapted already) practices, for example, pick glass up, move toward dishwasher, open dishwasher, and so forth, and must take in the right succession of activities to perform. The entire arrangement itself is either prompted through human demand by means of discourse handling or learned through perception of the task finished by a human demonstrator. Different cases of high-level state learning incorporate learning arrangements of known conduct for route through impersonation of a more proficient robot or human and learning and sequencing of primitive movements for full body movement in humanoid robots[16].

# CHAPTER 3. MATHEMATICAL PRELIMINARIES

## 3.1 Differential Dynamic Programming

Differential dynamic programming (DDP) is a second-order local trajectory optimization method which produces locally optimal trajectory and approximate local value function. DDP was proposed in 1966, but as applying DDP to constrained problems is difficult, it had not been introduced as a common approach to solve trajectory optimization problems. Recently due to its significant benefits, this technique is considered as one of the most important trajectory optimizers[25]–[27].

Actually in dynamic programming, a value function is used to produce locally optimal trajectories. A value function is sum of the accumulated future cost and the terminal cost given the current policy. DDP uses several iterations to solve nonlinear optimal control problems and each iteration of DDP includes backward path and forward path that is shown in Figure 2. In backward path, we compute optimal control policy of quadratic system expanded around reference trajectory, and in forward path we update reference trajectory by executing obtained optimal control policy[28], [29].

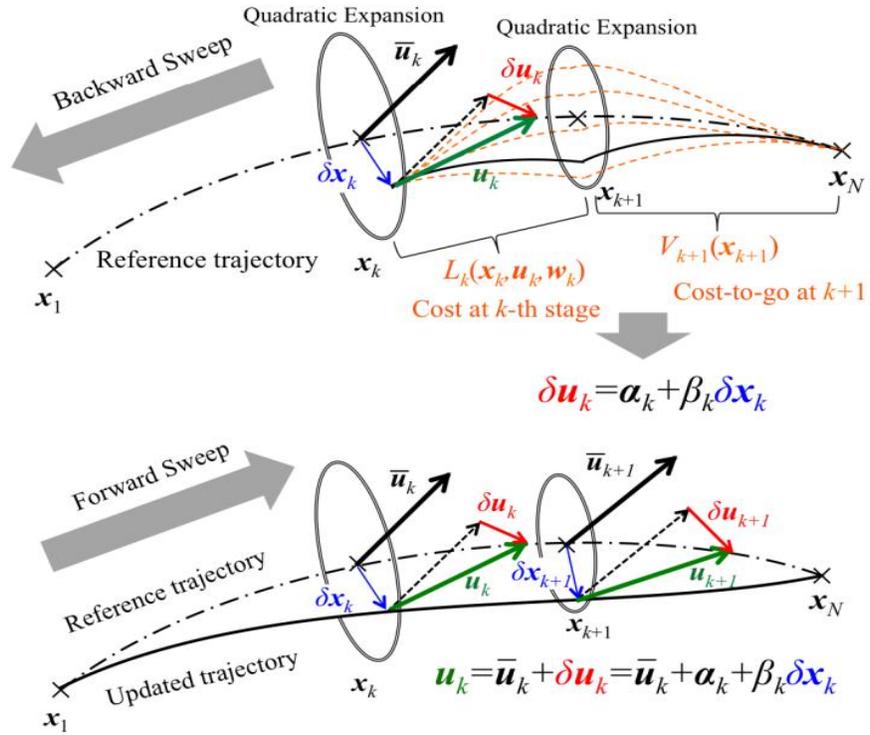

**Figure 2 : Backward sweep and forward sweep step in DDP[28]**

Discrete-time dynamical system and control policy can be defined as:

$$x_{k+1} = f_k(x_k, u_k; t_k) \qquad\qquad \textbf{2}$$

$$\{u_k^*\} := \{u_1^*, u_2^*, \dots, u_N^*\}$$

where $x_k \in \mathbb{R}^n$ is state at discretized time $k \in \{1, 2, \dots, N+1\}$, $u_k \in \mathbb{R}^m$ is control policy, $t_k \in \mathbb{R}$ is time, and $f_k: \mathbb{R}^n \times \mathbb{R}^m \times \mathbb{R} \to \mathbb{R}^n$ is the nonlinear function to define dynamics of the system.

Goal of the problem is finding the optimal control policy such that minimizes the objective function that is defined as sum of the accumulated cost over time and the terminal cost:

$$J(\{u_k^*\}) := h_{N+1}(x_{N+1}; t_{N+1}) + \sum_{k=1}^{N} l_k(x_k, u_k; t_k) \qquad \textbf{3}$$

here $h_{N+1}$ is terminal cost, and $l_k$ is cost at time $k$.

In dynamic programming we obtain the control input $u_k$ such that minimizing cost-to-go function $V_k$:

$$V_k(x_k) := h_{N+1}(x_{N+1}; t_{N+1}) + \sum_{i=k}^{N} l_i(x_i, u_i; t_i) \qquad \textbf{4}$$

Then we need to solve a recursive optimization problem to optimize the cost at time $k$ and the rest of cost-to-go function:

$$V_k^*(x_k) := \min[l_k(x_k, u_k; t_k) + V_{k+1}^*(x_{k+1})] \qquad \textbf{5}$$

Consider $V_k^*$ as optimal cost-to-go function.

Solving this problem by dynamic programming will lead to curse of dimensionality; however, DDP is based on quadratic expansions of Bellman's principle of optimality around a reference trajectory. If we consider $\bar{x}_k$ as reference trajectory, the quadratic expansion of Eq. (5) by considering $x_k(=\bar{x}_k + \delta x_k)$ and $u_k(=\bar{u}_k + \delta u_k)$ would be:

$$V_k^*(x_k) \approx V_k^*(\bar{x}_k) + V_x^{*(k)}\delta x_k + \frac{1}{2}\delta x_k^T V_{xx}^{*(k)}\delta x_k \qquad \textbf{6}$$

$$l_k(x_k, u_k) \approx l_k(\bar{x}_k, \bar{u}_k) + l_x^{(k)}\delta x_k + l_u^{(k)}\delta u_k + \frac{1}{2}\delta x_k^T l_{xx}^{(k)}\delta x_k + \delta x_k^T l_{xu}^{(k)}\delta u_k + \frac{1}{2}\delta u_k^T l_{uu}^{(k)}\delta u_k \quad \textbf{7}$$

$$V_{k+1}^*(x_{k+1}) \approx V_{k+1}^*(\bar{x}_{k+1}) + V_x^{*(k+1)}\delta x_{k+1} + \frac{1}{2}\delta x_{k+1}^T V_{xx}^{*(k+1)}\delta x_{k+1} \qquad \textbf{8}$$

here $k$ represents time and other subscripts means partial derivative with respect to $x_k$ or $u_k$. Using the Eq.(1) we can find $\delta \mathrm{x}_{k+1}$ as:

$$\delta \mathrm{x}_{k+1} \approx f_x^{(k)} \delta \mathrm{x}_k + f_u^{(k)} \delta \mathrm{u}_k + \frac{1}{2} \delta x_k^T * f_{xx}^{(k)} \delta x_k + \delta x_k^T * f_{xu}^{(k)} \delta \mathrm{u}_k + \frac{1}{2} \delta u_k^T * f_{uu}^{(k)} \delta \mathrm{u}_k \quad 9$$

where the operator * is defined as $(A * B)_{jk} = A_i B_{ijk}$, where the subscripts mean the element in tensor note. Therefore, the Eq. (5) can be described as:

$$l_k(x_k, u_k; t_k) + V_{k+1}^*(x_{k+1}) = Q_0 + [q_x^T \, q_u^T] \begin{bmatrix} \delta \mathrm{x}_k \\ \delta \mathrm{u}_k \end{bmatrix} + \frac{1}{2} \left[ \delta x_k^T \, \delta u_k^T \right] \begin{bmatrix} Q_{xx} & Q_{xu} \\ Q_{xu} & Q_{uu} \end{bmatrix} \begin{bmatrix} \delta \mathrm{x}_k \\ \delta \mathrm{u}_k \end{bmatrix} \quad 10$$

Let us assume $Q_{uu}$ is a positive definite matrix; the optimal control variations can be obtained as stationary points of Eq.(10):

$$\delta u_k = \alpha_k + \beta_k \delta \mathrm{x}_k \quad\quad\quad\quad 11$$

where:

$$\alpha_k := -Q_{uu}^{-1} q_u \quad\quad\quad\quad 12$$

and

$$\beta_k := -Q_{uu}^{-1} Q_{xu}^T \quad\quad\quad\quad 13$$

DDP computes the optimal control policy $\alpha_k$ and $\beta_k$ to minimize $l_k(x_k, u_k; t_k) + V_{k+1}^*(x_{k+1})$ in the backward path. And in forward path, we update reference trajectory by using obtained optimal control such that $u_k = \bar{u}_k + \delta \mathrm{u}_k = \bar{u}_k + \alpha_k + \beta_k \delta \mathrm{x}_k$[25], [28], [30], [31].

## 3.2 Stochastic Differential Dynamic Programming

We define the cost function of nonlinear stochastic continuous optimal control problems as:

$$v^u(x,t) = E[h(x(T)) + \int_{t_0}^T l(\tau, x(\tau), u(\tau, x(\tau)))] \qquad 14$$

And stochastic dynamics of the system can be defined as:

$$dx = f(x,u)dt + F(x,u)d\omega \qquad 15$$

where $x$ is state, $u$ is the control policy and $d\omega$ is Brownian noise. $l$ is the cost rate which is a function of state and control policy and $h$ is the terminal cost. The cost-to-go $v$ is defined as the expected cost accumulated over time horizon $t_o, ..., T$ starting from initial state to the final state. To define the derivatives clearly we express dynamics as a function $\Phi$ of the states, control and noise.

$$\Phi(x,u,d\omega) = f(x,u)dt + F(x,u)d\omega \qquad 16$$

Given a reference trajectory of states and actions $(\bar{x}, \bar{u})$, we expand the dynamics around reference trajectory to second order:

$$\Phi(\bar{x} + \delta x, \bar{u} + \delta u, d\omega) = \Phi(\bar{x}, \bar{u}, d\omega) + \Phi_x . \delta x + \Phi_u . \delta u + O(\delta x, \delta u, \delta \omega) \qquad 17$$

where:

$$O^{(j)}(x, \delta u, \delta \omega) = \frac{1}{2} [\delta x \, \delta u] \begin{bmatrix} \Phi_{xx}^j & \Phi_{xu}^j \\ \Phi_{xu}^j & \Phi_{uu}^j \end{bmatrix} \begin{bmatrix} \delta x \\ \delta u \end{bmatrix} \qquad 18$$

$$\Phi_x = \nabla_x f(x, u)\delta t + \nabla_x(\sum_{i=1}^m F_c^i d\omega_t^{(i)}) \qquad \mathbf{19}$$

$$\Phi_u = \nabla_u f(x, u)\delta t + \nabla_u(\sum_{i=1}^m F_c^i d\omega_t^{(i)}) \qquad \mathbf{20}$$

$$\Phi_{xx} = \nabla_{xx} f(x, u)\delta t + \nabla_{xx}(\sum_{i=1}^m F_c^i d\omega_t^{(i)}) \qquad \mathbf{21}$$

$$\Phi_{uu} = \nabla_{uu} f(x, u)\delta t + \nabla_{uu}(\sum_{i=1}^m F_c^i d\omega_t^{(i)}) \qquad \mathbf{22}$$

$$\Phi_{xu} = \nabla_{xu} f(x, u)\delta t + \nabla_{xu}(\sum_{i=1}^m F_c^i d\omega_t^{(i)}) \qquad \mathbf{23}$$

$$\Phi_{ux} = \nabla_{ux} f(x, u)\delta t + \nabla_{ux}(\sum_{i=1}^m F_c^i d\omega_t^{(i)}) \qquad \mathbf{24}$$

After expanding the dynamics up to second order we can transition from continuous to discrete time:

$$\delta x_{t+\delta t} = (A_t \delta x_t + B_t \delta u_t + \Gamma_t \xi + O_d) \qquad \mathbf{25}$$

where:

$$A_t = I_{n \times n} + \nabla_x f(x, u)\delta t \qquad \mathbf{26}$$

$$B_t = \nabla_u f(x, u)\delta t \qquad \mathbf{27}$$

$$\Gamma_t = \nabla_x F(x, u).\delta x + \nabla_u F(x, u).\delta u + F(x, u) \qquad \mathbf{28}$$

and $\xi$ is the random variable.

Substitution of the discretized dynamics in the second-order value function expansion results in [32], [33]:

$$V(\overline{x}_{t+\delta t} + \delta x_{t+\delta t}) = V(\overline{x}_{t+\delta t})$$

$$+ V_x^T(A_t\delta x_t + B_t\delta u_t + \Gamma_t\xi + O_d)$$

$$+ (A_t\delta x_t + B_t\delta u_t + \Gamma_t\xi + O_d)^T$$

$$\times V_{xx}(A_t\delta x_t + B_t\delta u_t + \Gamma_t\xi + O_d) \qquad \mathbf{29}$$

# CHAPTER 4. USING APPROXIMATE MODELS

An accurate dynamics model is not always needed to find appropriate controllers whereas a more accurate dynamics model will normally only develop the controllers when we use model-based reinforcement learning (RL) algorithms. For example, when a human learns to drive a car, he can learn to drive the car reliably without essentially building up an accurate dynamics model. Instead, we can learn to drive a car using a crude model and a few real-life trials. Here we present an algorithm that requires only an approximate model and only a small number of real-life trials. The main idea is that a real-life trial together with an approximate model can be sufficient to obtain reasonable policy gradient estimates.

## 4.1 Introduction

In model-based reinforcement learning, a model first is constructed for the real system and extracts the control policy that is optimal in this model. Then this control policy is deployed in the real system. Recent research in reinforcement learning and optimal control has generated effective algorithms to derive optimal control policies for a huge range of models and reward functions[2], [3], [34], [35].

But, for a lot of main control problems, specifically tasks with high-dimensional and continuous states and policy, it is very difficult to construct a perfect model of the Markov decision process. When we acquire a policy using an inexact model, we frequently find a policy that does not work well in real life and it only has good acceptable result in simulation (i.e., the policy works well in the predefined model). On the other hand, in model-free policy search, there is the other side that is searching for control policy only on the real system, without ever explicitly constructing a model. This approach is successfully applied to a limited number of applications[36]; these model-free RL approaches need enormous, and impossibly large, numbers of real-life experiments.

The huge number of real-life experiments required by model-free RL is in serious contrast to humans' method of learning to do a complicated task. For instance, when a driving student wants to learn to drive a car through a 90-degree turn, particularly learning the amount of steering required, on the first attempt he may take the turn short. He will then change and take the next turn wider (or possibly too wide). Naturally, it will need only a few attempts to learn to perform the turn appropriately.

The driving student obviously does not have an accurate model of the car system. Neither does he require a huge number of real-life experiments. However, he combines a simple model of the car with a few number of real-life attempts to learn to do appropriately.

Here we describe the idea and develop an algorithm that exploits an approximate model to rapidly learn to do well on real systems. The main idea is to use a real-life trial to assess a control policy and then use the model to estimate the derivative of the evaluation with respect to the policy parameters and propose local modifications. For instance, in a car example if the current policy takes the turn too short, then driving in real life will be indicating that the policy is driving the turn too short. Conversely, even a very inaccurate model of the car can indicate the change we should make is to turn the steering wheel clockwise (instead of counter-clockwise) to modify for this error. Specifically, we do not require extra real-life experiments of turning the steering wheel both clockwise and counter-clockwise to understand the correct direction. So, even an approximate model of the car helps to significantly decrease the amount of real-life attempts required compared to model-free algorithms that run directly on the real system.

Compared to typical model-based algorithms, this approach has the benefit that it does not need a precise model of the Markov decision process. In spite of the improvement in learning algorithms for constructing well dynamics models, it is still an extremely complicated problem to model the comprehensive dynamics of various systems, such as helicopters, cars and particular aircraft. This method requires only an approximate model of the system.

While an extension to stochastic systems is imaginable, the algorithm presented in this chapter only applies to systems that are close to deterministic. Here we assume a continuous state and action space[10].

## 4.2 Algorithm

The algorithm needs inputs that are an approximate MDP $\hat{M} = (S, A, \hat{T}, H, s_0, R)$ and a local policy improvement algorithm such as differential dynamic programming. The MDP dynamics model $\hat{T}$ is an approximate model of the true dynamics $T$ of the true MDP $M = (S, A, M, H, s_0, R)$.

The local policy improvement algorithm could be any technique that iteratively improves the current policy locally. We use the most efficient one, which is differential dynamic programming.

The algorithm continues as:

1. Set $i = 0$. Set the initial model approximate $\hat{T}^{(0)} = \hat{T}$.

2. Running the DDP to find the locally optimal control policy $\pi_{\theta^{(0)}}$ for the MDP $\hat{M}^{(0)} = (S, A, \hat{T}^{(0)}, H, s_0, R)$.

3. Implement the obtained policy $\pi_{\theta^{(i)}}$ in the real system and record the resulting trajectory $s_0^{(i)}, a_0^{(i)}, s_1^{(i)}, a_1^{(i)}, ..., s_H^{(i)}, a_H^{(i)}$.

4. Update the model $\hat{T}^{(i+1)}$ by adding a time-dependent bias term to the original model $\hat{T}$. In more detail, set $\hat{f}_t^{(i+1)}(s, a) = \hat{f}_t(s, a) + s_{t+1}^{(i)} - \hat{f}_t\left(s_t^{(i)}, a_t^{(i)}\right)$ for all times $t$.

5. Use DDP in the MDP $\hat{M}^{(i+1)} = (S, A, \hat{T}^{(i+1)}, H, s_0, R)$ to find a local policy improvement direction $d^{(i)}$ such that $\hat{U}\left(\pi_{\theta^{(i)}+\alpha d^{(i)}}\right) \geq \hat{U}(\pi_{\theta^{(i)}})$ for some step size $\alpha > 0$. Here $\hat{U}$ is the utility as evaluated in $\hat{M}^{(i+1)}$.

6. Use a line search to acquire the improved policy $\pi_{\theta^{(i+1)}}$, where $\theta^{(i)} + \alpha^{(i)} d^{(i)}$. During the line search, assess the policies $\pi_{\theta^{(i)}+\alpha d^{(i)}}$ in the real system.

7. Algorithm will terminate if in the $n$ previous iterations the line search found an improved policy such that $\hat{U}\left(\pi_{\theta^{(i)}+\alpha d^{(i)}}\right) - \hat{U}(\pi_{\theta^{(i)}}) < \varepsilon$, else increase the iteration number $i = i + 1$ and move to step 3.

The algorithm is started with the control policy $\pi_{\theta^{(0)}}$, which is locally optimal for the first approximate dynamics model. In following iterations, results in the real system progress, and consequently the updated control policy works at least good enough like the model-based policy $\pi_{\theta^{(0)}}$, and feasibly better.

A time-dependent bias to the model is added in each iteration $i$: the term $s_{t+1}^{(i)} - \hat{f}_t\left(s_t^{(i)}, a_t^{(i)}\right)$ in step 4 of the algorithm. In the updated model $\hat{T}^{(i+1)}$ the model is modified such that $\hat{f}_t^{(i+1)}\left(s_t^{(i)}, a_t^{(i)}\right) = s_{t+1}^{(i)}$ for all times $t$. Therefore, the updated model $\hat{T}^{(i+1)}$ precisely calculates the real system trajectory $s_0^{(i)}, s_1^{(i)}, \ldots, s_H^{(i)}$ acquired when implementing the control policy $\pi_{\theta^{(i)}}$.

Hence, after calculating the improvement direction in step 5, the algorithm calculates the derivatives along the correct state-action trajectory. However, the model-based approach calculates the derivatives along the state-action trajectory predicted by the predefined model $\hat{T}$, which would not match to the true state-action trajectory when the predefined model is an inexact model[11].

# CHAPTER 5. EXPERIMENTS AND RESULTS

## 5.1 Introduction

We implemented the algorithm in environment of Spumone, a video game based on mechanical engineering concepts(Figure 3).

Spumone is a video game framework designed for learning engineering by Brianno Coller in Northern Illinois University. The goal is to make some of the hard work in learning fundamental engineering science concepts and problem solving techniques much more engaging and effective than traditional textbook homework problems.

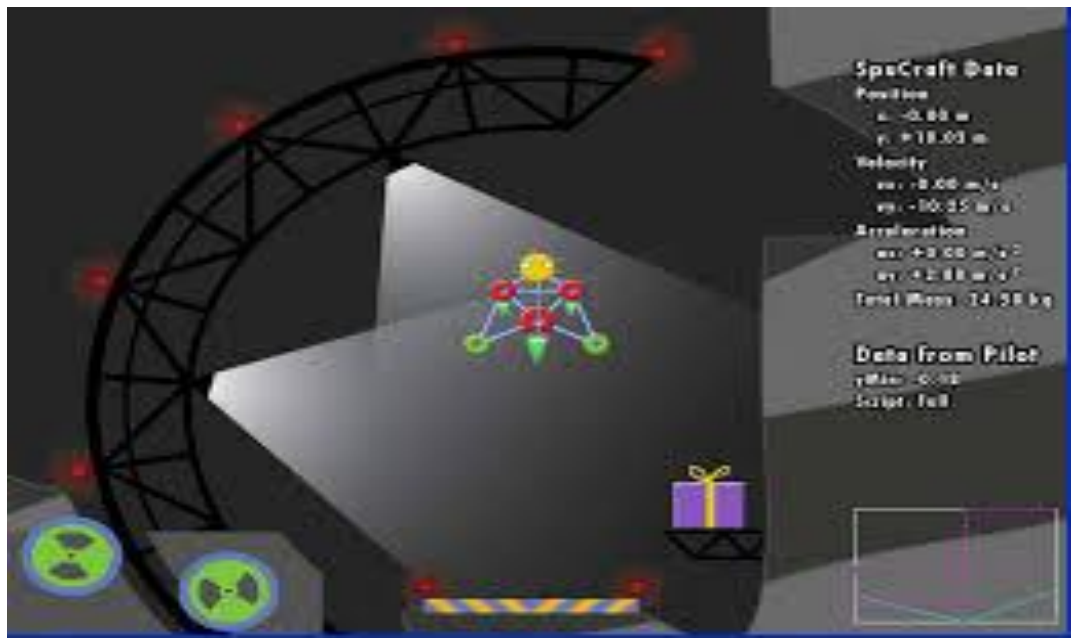

**Figure 3 : Spumone screen shot**

Rollo World is an environment defined in Spumone that is a good place to practice applying rigid body kinematics relationships. In Figure 4 the schematic of the Rollo craft is presented.

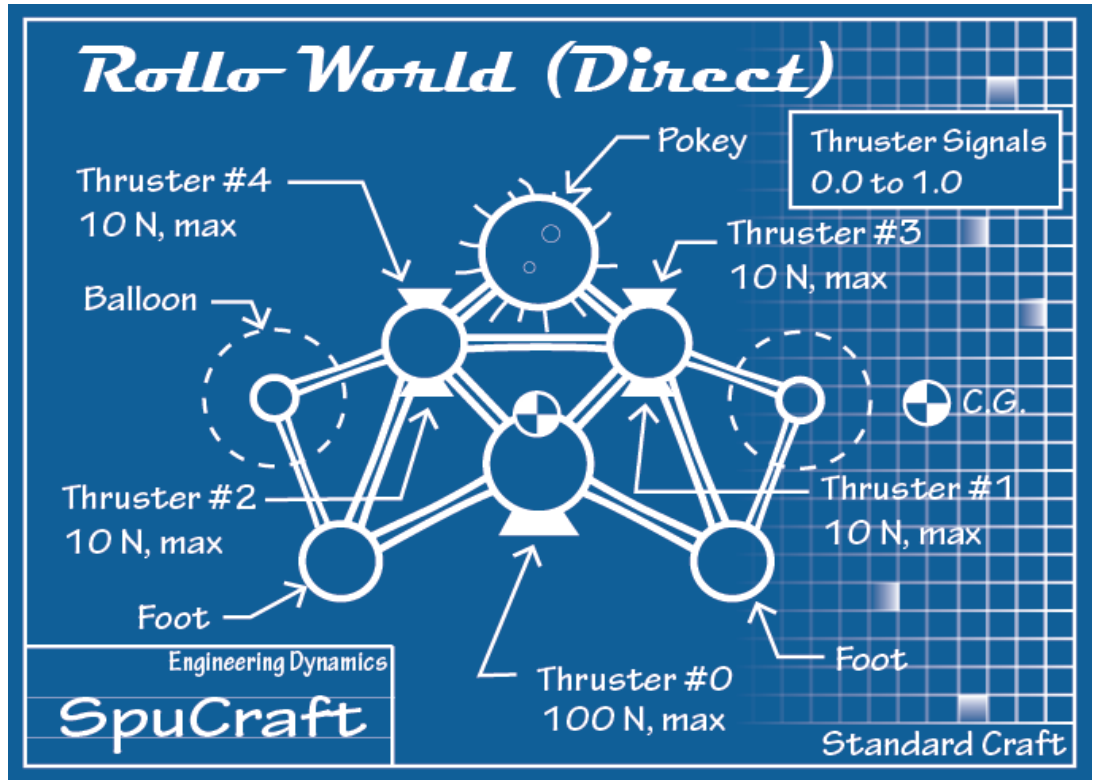

**Figure 4: Spumone craft schematic**

First we defined a trajectory manually to run the Spumone craft from the start point to the landing point(Figure 5). In this step we only define the states that we want to reach and did not consider dynamics of system and the possibility of the trajectory due to its dynamics.

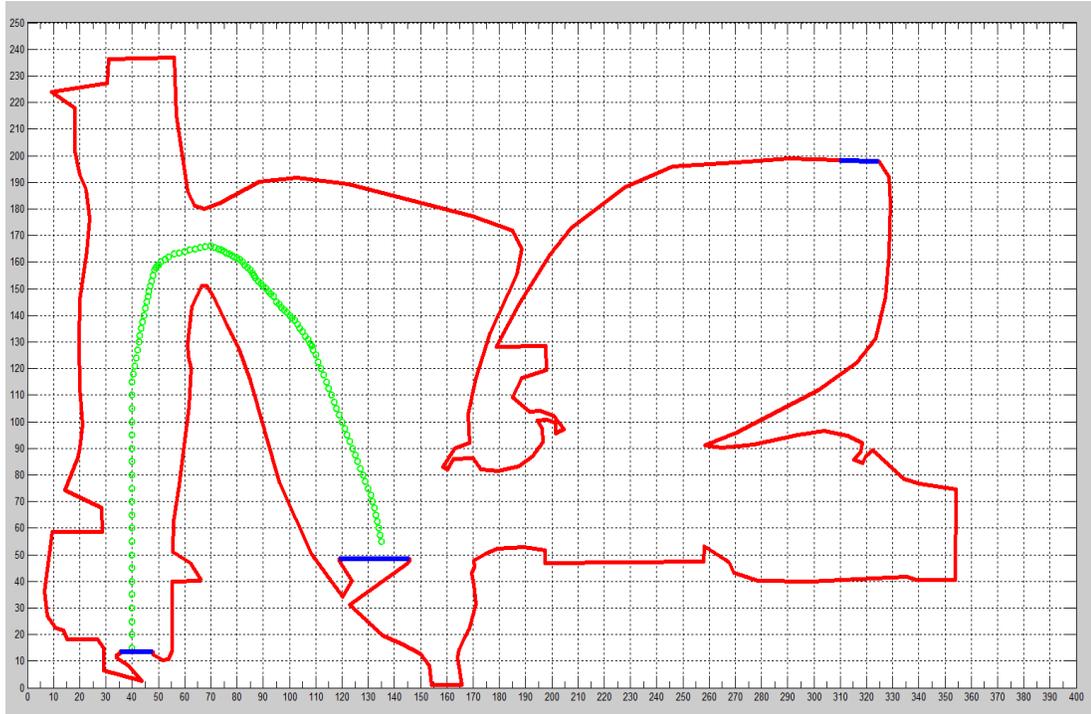

**Figure 5: Wide view of Spumone deck and the reference trajectory**

As the local trajectory optimizer we used the DDP to find the optimal control policy to follow predefined reference trajectory as close as it is possible due to dynamics of system. In this regard, defining the suitable cost function plays a very important role to achieve the required goals. Setting coefficient for each element of cost function with respect to each other requires trial and error and it would be more complicated when you want to achieve different goals that have contrast such as trajectory following, limits on controls, smoothness of the trajectory, safe landing, and minimization of controls. All steps of the algorithm to find the optimal control policy are modeled in MATLAB R2012a and exported to DAT file from MATLAB and transferred to be used instead of joystick inputs in Spumone.

## 5.2 Results Evaluation

In this section, we describe experiments that were conducted with two different methods to evaluate the performance of the algorithm: (1) running the Spumone using the algorithm control policy as inputs and (2) running Spumone using human expert input by joystick. Following criteria are used to evaluate the performance of trajectory achieved by the presented algorithm and trajectory of the human expert.

### 5.2.1 Errors

First we should check whether a trajectory contains errors making the trajectory fundamentally impractical and consequently unacceptable. These errors discover actions that would never occur in a real system given the aforementioned task and conditions. (1) A stop error is generated when the agent has totally stopped throughout its trajectory (when velocity is 0 between the start and goal). (2) A collision error is generated when the craft has collided with the deck.

According to Figure 6, although velocity in x and y directions both algorithm and expert trajectory experience the zero, the total velocity never reaches zero between the start point and destination. So in this criteria both algorithm and expert trajectory performed well.

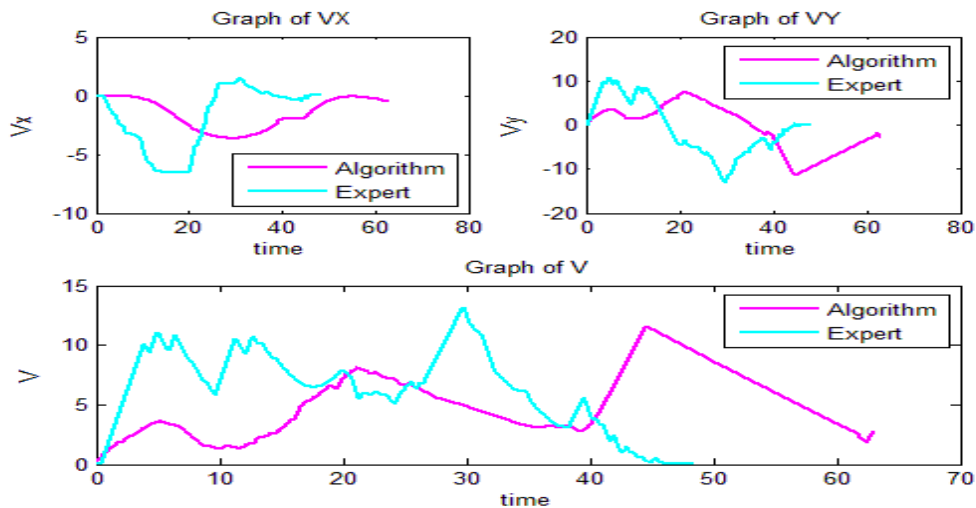

**Figure 6: Velocity in x and y directions and total velocity magnitude**

As demonstrated in Figure 7 fortunately neither algorithm trajectory nor expert trajectory has collision with the deck.

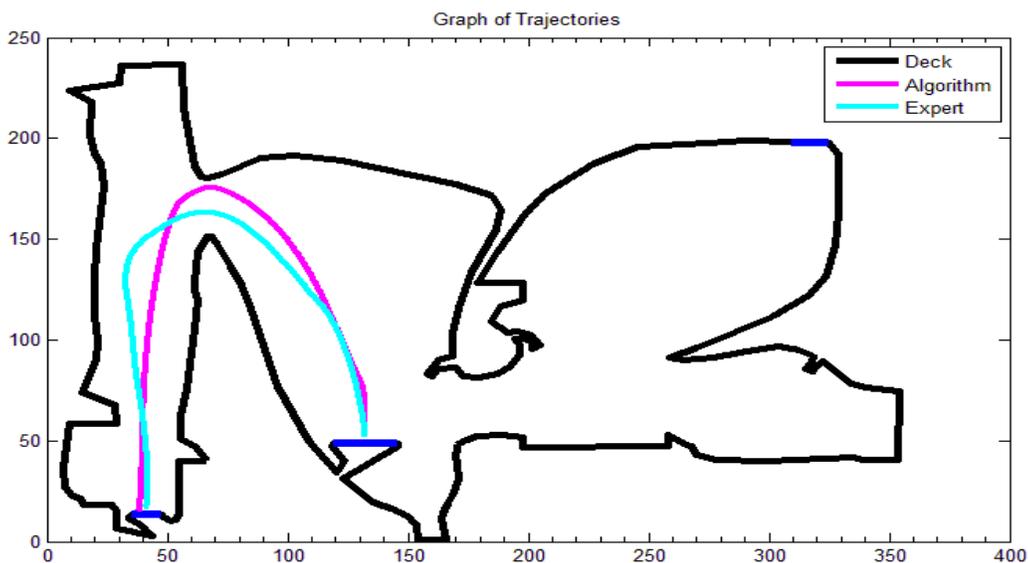

**Figure 7: Spumone Deck and trajectories resulted by algorithm and expert**

### 5.2.2 Deviation from Reference Trajectory

One of most important factors for trajectory following task is evaluating the deviation from the reference trajectory that is illustrated in Figure 8. As our reference trajectory is designed intuitively and there is no possibility of assurance for it, both trajectories expected to have considerable deviation from the reference trajectory due to dynamics of the system, but as demonstrated in Figure 8, both are far enough from the deck to prevent collision.

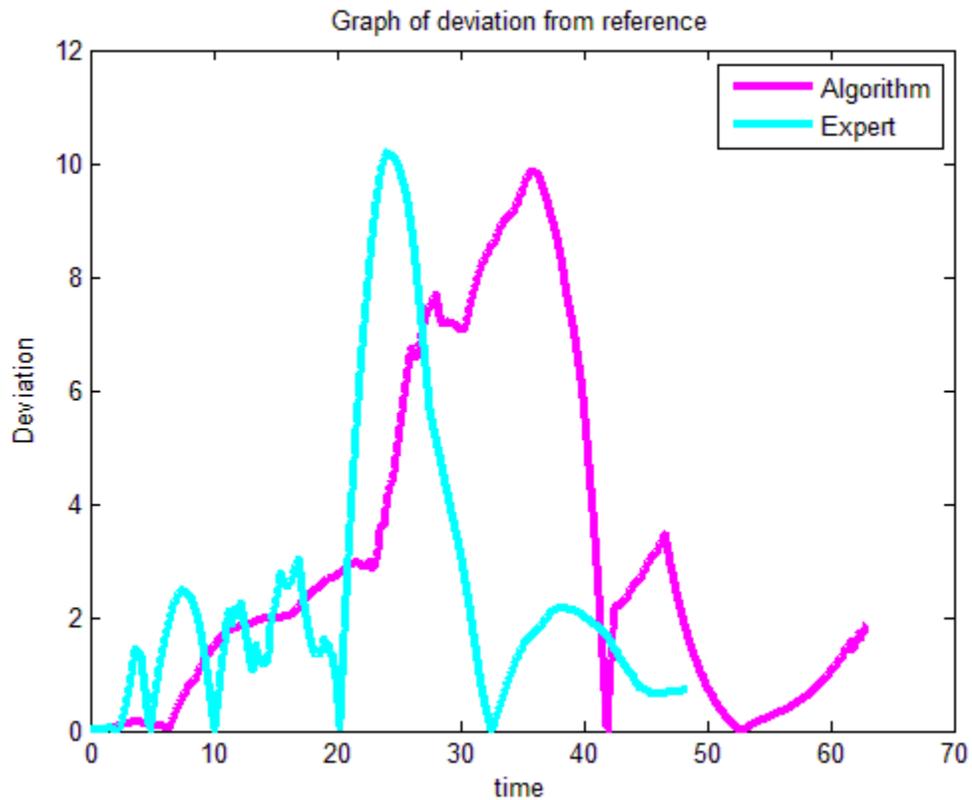

**Figure 8: Deviation from the reference trajectory**

### 5.2.3 Smoothness

In this part we evaluate the smoothness of the trajectory. This criterion is inspired by studies which showed that human trajectories maximized smoothness and that smoothness maximization could be achieved by jerk minimization. Smoothness is therefore evaluated by computing the mean jerk amplitude of the trajectory. Smaller values mean smoother trajectories. As shown in Figure 9 trajectory obtained by the presented algorithm is smoother than the expert trajectory.

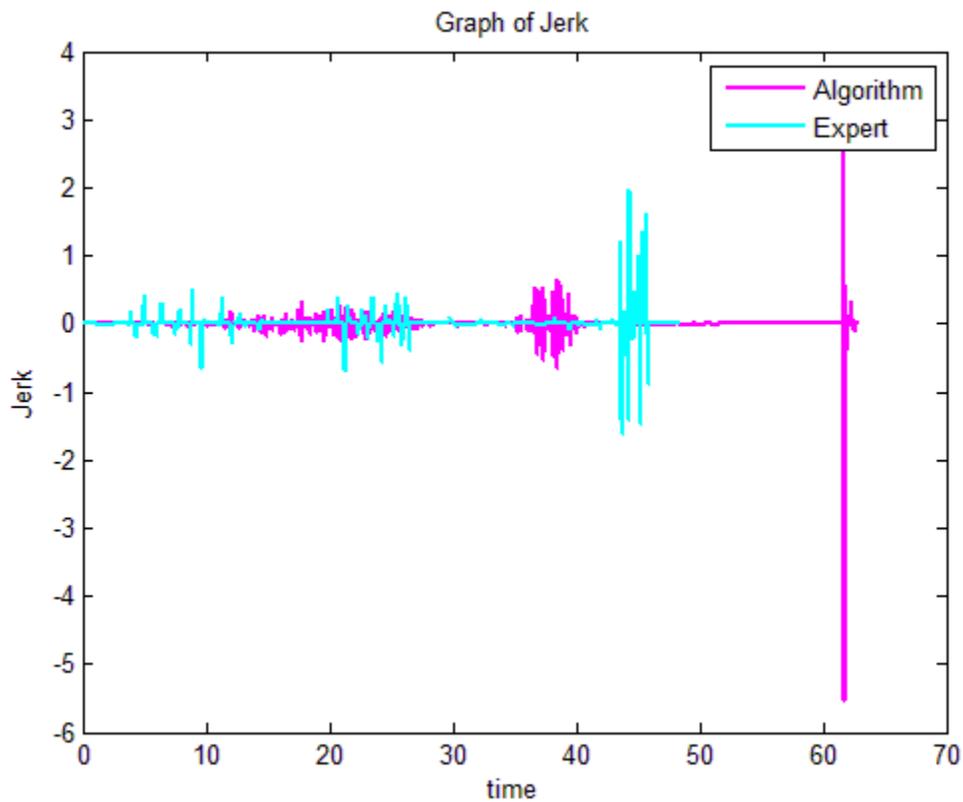

**Figure 9: Plot of jerk over time**

### 5.2.4 Energy Optimization (Mechanical Work)

One of the most important factors in each evaluation is comparing the required energy for performing the trajectory. In Figure10 and 11 we compare the mechanical work in each time step of both trajectories and we also compute the total required energy for running each control policy.

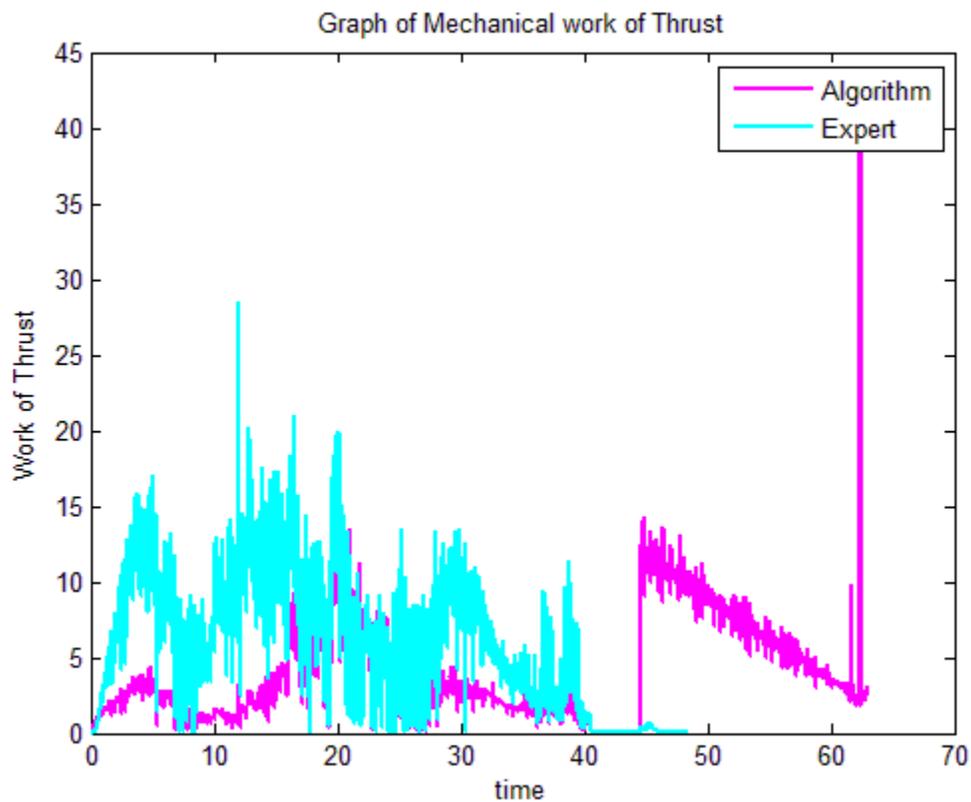

**Figure 10: Mechanical work of thrust**

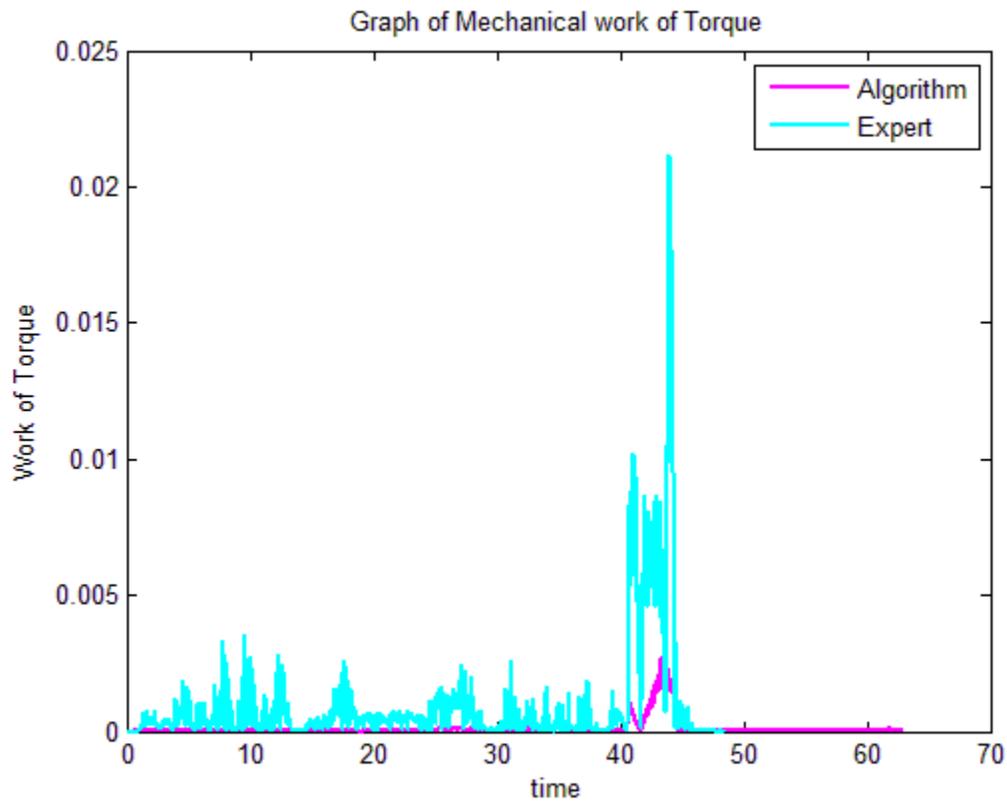

**Figure 11: Mechanical work of torque**

### 5.2.5 Duration

Evaluating duration and length of the trajectories is another factor that is considered here. In both sub-criteria (duration and length), human expert has better performance compared with the algorithm control policy. Duration of human expert trajectory is about 48.4 seconds and the required time to execute the optimal control policy obtained by algorithm is 63 seconds. Total length of the trajectory of human expert is 303.1 meters and the total length of the trajectory by algorithm is 309.9 meters.

### 5.2.6 Rotational Movement Parameters

This factor evaluates the difference between both the angle and angular velocity of the algorithm trajectory and the angular velocity of an expert trajectory.

As demonstrated in Figure 12 and 13, it is obvious that both angle of the craft and the rotational velocity of the craft of the human expert trajectory have more fluctuations compared with trajectory by algorithm that has very little and smooth variation.

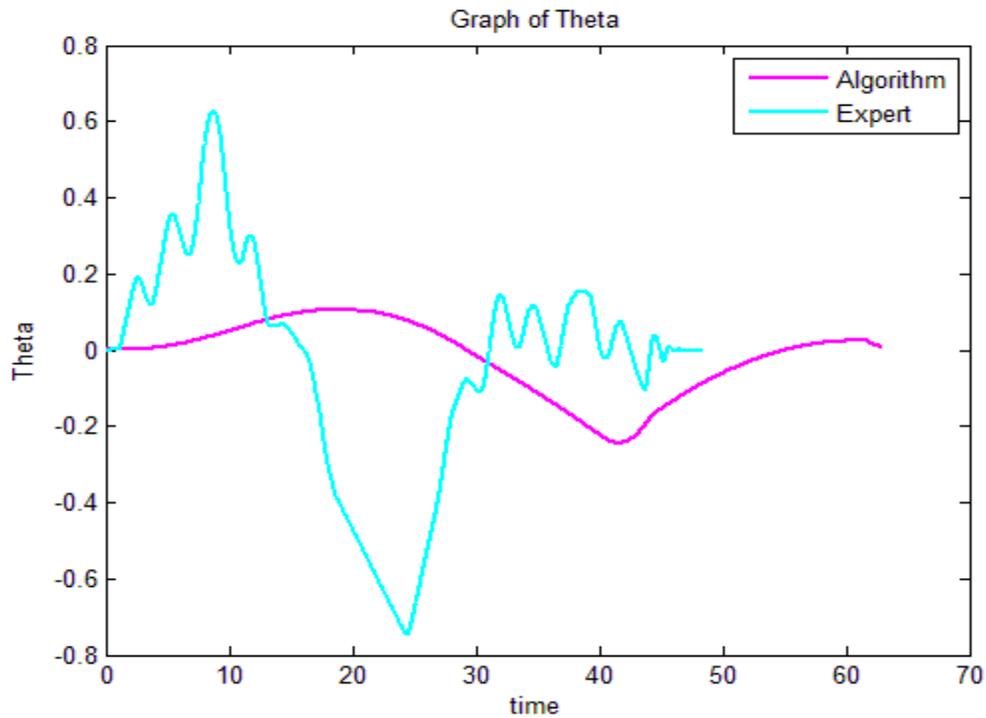

**Figure 12: Theta over time**

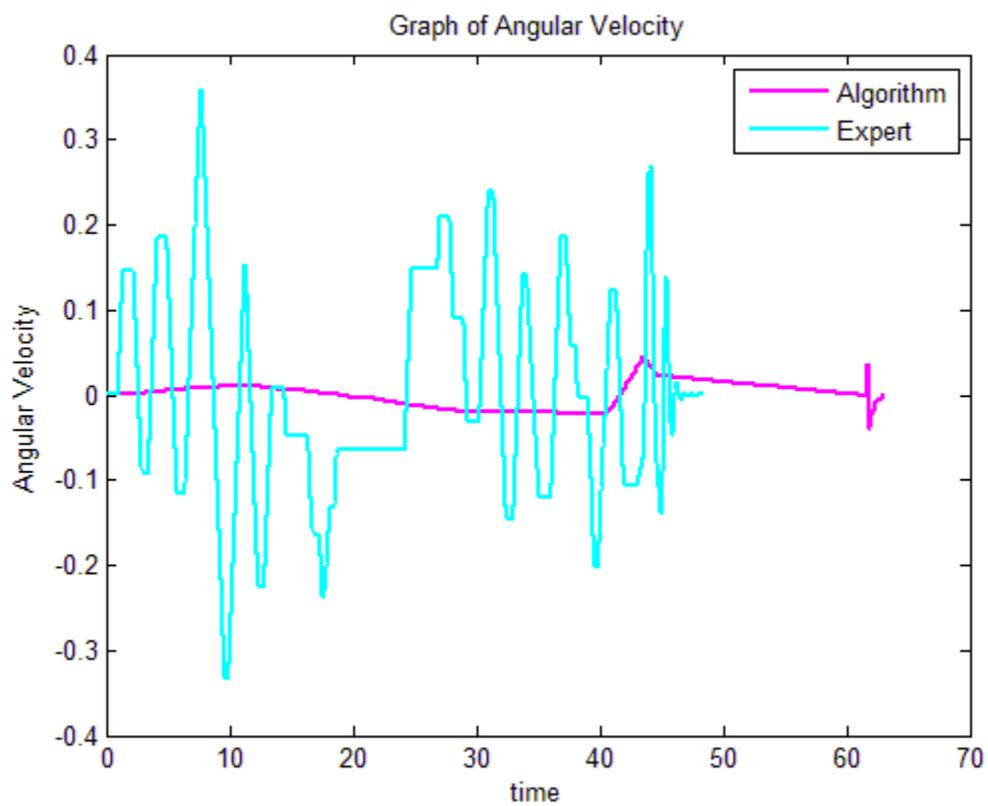

**Figure 13: Omega over time**

## CHAPTER 6. CONCLUSION AND FUTURE WORKS

Implementation of traditional control and model-based reinforcement learning resulted in some fundamental difficulties. Control task needs to be defined in a compatible form with control algorithm; also, dynamics model of the system is required and finally, beside these issues, finding good closed-loop controllers for these problems often is computationally expensive.

An algorithm is developed and implemented to use an approximate model and a limited number of real-life experiments to calculate a control policy that works well in real systems. Our experiment with Spumone (a video game) shows that the algorithm can generally perform at least as well as a human expert. Even from some important point of view such as mechanical work to perform the task or smoothness, it significantly outperformed the human expert.

Most interesting and practical direction for this research is to develop a stochastic extension for this algorithm that can perform well in stochastic environments such as wind.

Due to difference between human abilities area and computer abilities area, another interesting future of research would be distinguishing the types of problems and tasks in which algorithm can outperform the human and vice versa and finally combine both benefits.

# REFERENCES

[1]     R. S. Sutton, A. G. Barto, and R. J. Williams, "Reinforcement learning is direct adaptive optimal control," *Control Syst. IEEE*, vol. 12, no. 2, pp. 19–22, 1992.

[2]     R. S. Sutton and A. G. Barto, *Reinforcement learning: An introduction*, vol. 1, no. 1. MIT Press Cambridge, 1998.

[3]     L. P. Kaelbling, M. L. Littman, and A. W. Moore, "Reinforcement learning: A survey," *J. Artif. Intell. Res.*, pp. 237–285, 1996.

[4]     J. Canny and J. Reif, "New lower bound techniques for robot motion planning problems," in *Foundations of Computer Science, 1987., 28th Annual Symposium on*, 1987, pp. 49–60.

[5]     R. A. Brooks, "A robot that walks; emergent behaviors from a carefully evolved network," *Neural Comput.*, vol. 1, no. 2, pp. 253–262, 1989.

[6]     P. Maes, "The agent network architecture (ANA)," *Acm sigart Bull.*, vol. 2, no. 4, pp. 115–120, 1991.

[7]     B. Kuipers and Y.-T. Byun, "A robot exploration and mapping strategy based on a semantic hierarchy of spatial representations," *Rob. Auton. Syst.*, vol. 8, no. 1, pp. 47–63, 1991.

[8]     S. Mahadevan and J. Connell, "Automatic programming of behavior-based robots using reinforcement learning," *Artif. Intell.*, vol. 55, no. 2, pp. 311–365, 1992.

[9]     A. W. Moore and T. Hall, "Efficient memory-based learning for robot control," 1990.

[10]    P. Abbeel, M. Quigley, and A. Y. Ng, "Using inaccurate models in reinforcement learning," in *Proceedings of the 23rd international conference on Machine learning*, 2006, pp. 1–8.

[11]    P. Abbeel, *Apprenticeship learning and reinforcement learning with application to robotic control*. ProQuest, 2008.

[12]    A. Billard, S. Calinon, Rǐ. Dillmann, and S. Schaal, "Robot programming by

demonstration," in *Springer handbook of robotics*, Springer, 2008, pp. 1371–1394.

[13]   A. Billard, S. Calinon, R. Dillmann, and S. Schaal, "Survey: Robot programming by demonstration," MIT Press, 2008.

[14]   B. D. Argall, S. Chernova, M. Veloso, and B. Browning, "A survey of robot learning from demonstration," *Rob. Auton. Syst.*, vol. 57, no. 5, pp. 469–483, 2009.

[15]   S. Schaal, A. Ijspeert, and A. Billard, "Computational approaches to motor learning by imitation," *Philos. Trans. R. Soc. B Biol. Sci.*, vol. 358, no. 1431, pp. 537–547, 2003.

[16]   A. Billard and M. Arbib, *Imitation*, MIT Press, 2002.

[17]   C. L. Nehaniv and K. Dautenhahn, "Like me?-measures of correspondence and imitation," *Cybern. Syst.*, vol. 32, no. 1–2, pp. 11–51, 2001.

[18]   C. Breazeal, A. Brooks, J. Gray, G. Hoffman, C. Kidd, H. Lee, J. Lieberman, A. Lockerd, and D. Mulanda, "Humanoid robots as cooperative partners for people," *Int. J. Humanoid Robot.*, vol. 1, no. 2, pp. 1–34, 2004.

[19]   C. L. Nehaniv and K. Dautenhahn, *Imitation and social learning in robots, humans and animals: behavioural, social and communicative dimensions*. Cambridge University Press, 2007.

[20]   P. Kormushev, S. Calinon, and D. G. Caldwell, "Imitation learning of positional and force skills demonstrated via kinesthetic teaching and haptic input," *Adv. Robot.*, vol. 25, no. 5, pp. 581–603, 2011.

[21]   J. Nakanishi and S. Schaal, "Feedback error learning and nonlinear adaptive control," *Neural Networks*, vol. 17, no. 10, pp. 1453–1465, 2004.

[22]   D. Lee and C. Ott, "Incremental kinesthetic teaching of motion primitives using the motion refinement tube," *Auton. Robots*, vol. 31, no. 2–3, pp. 115–131, 2011.

[23]   K. Mülling, J. Kober, O. Kroemer, and J. Peters, "Learning to select and generalize striking movements in robot table tennis," *Int. J. Rob. Res.*, vol. 32, no. 3, pp. 263–279, 2013.

[24]   D. H. Grollman and O. C. Jenkins, "Incremental learning of subtasks from unsegmented demonstration," in *Intelligent Robots and Systems (IROS), 2010 IEEE/RSJ International Conference on*, 2010, pp. 261–266.

[25]   D. Jacobson and D. Mayne, "Differential dynamic programming," American Elsevier Publishing Company, New York, New York, 1970.

[26] G. Lantoine and R. P. Russell, "A Hybrid Differential Dynamic Programming Algorithm for Constrained Optimal Control Problems. Part 1: Theory," *J. Optim. Theory Appl.*, vol. 154, no. 2, pp. 382–417, 2012.

[27] G. J. Whiffen, "Static/dynamic control for optimizing a useful objective." Google Patents, 17-Dec-2002.

[28] N. Ozaki, S. Campagnola, C. H. Yam, and R. Funase, "Diffrential dynamic programming approach for robust-optimal low-thrust trajectory design considring uncertainty." 25th International Symposium on Space Flight Dynamics ISSFD, October, 2015, Munich, Germanys

[29] J. T. Betts, "Survey of numerical methods for trajectory optimization," *J. Guid. Control. Dyn.*, vol. 21, no. 2, pp. 193–207, 1998.

[30] Y. Tassa, N. Mansard, and E. Todorov, "Control-limited differential dynamic programming," in *2014 IEEE International Conference on Robotics and Automation (ICRA)*, 2014, pp. 1168–1175.

[31] E. Todorov and Y. Tassa, "Iterative local dynamic programming," in *2009 IEEE Symposium on Adaptive Dynamic Programming and Reinforcement Learning*, 2009, pp. 90–95.

[32] E. Theodorou, Y. Tassa, and E. Todorov, "Stochastic differential dynamic programming," in *American Control Conference*, 2010, pp. 1125–1132.

[33] E. Todorov and W. Li, "A generalized iterative LQG method for locally-optimal feedback control of constrained nonlinear stochastic systems," in *Proceedings of the 2005, American Control Conference, 2005.*, 2005, pp. 300–306.

[34] D. P. Bertsekas, D. P. Bertsekas, D. P. Bertsekas, and D. P. Bertsekas, *Dynamic programming and optimal control*, vol. 1, no. 2. Athena Scientific Belmont, MA, 1995.

[35] J. Abounadi, D. Bertsekas, and V. S. Borkar, "Learning algorithms for Markov decision processes with average cost," *SIAM J. Control Optim.*, vol. 40, no. 3, pp. 681–698, 2001.

[36] N. Kohl and P. Stone, "Policy gradient reinforcement learning for fast quadrupedal locomotion," in *Robotics and Automation, 2004. Proceedings. ICRA'04. 2004 IEEE International Conference on*, 2004, vol. 3, pp. 2619–2624.